\documentclass[letterpaper]{article} 
\usepackage{aaai25}  
\usepackage{times}  
\usepackage{helvet}  
\usepackage{courier}  
\usepackage[hyphens]{url}  
\usepackage{graphicx} 
\usepackage{natbib}  
\usepackage{caption} 
\usepackage{algorithm}
\usepackage{algorithmic}

\usepackage{newfloat}
\usepackage{listings}
\DeclareCaptionStyle{ruled}{labelfont=normalfont,labelsep=colon,strut=off} 
\floatstyle{ruled}
\newfloat{listing}{tb}{lst}{}
\floatname{listing}{Listing}
\usepackage[utf8]{inputenc}
\usepackage{amssymb}  
\usepackage{pifont}   
\usepackage{xcolor}   
\usepackage{multirow}
\usepackage{graphicx}
\makeatletter

\newcommand{\Rmnum}[1]{\expandafter\@slowromancap\romannumeral #1@}
\makeatother
\usepackage{booktabs}
\usepackage{appendix}

\title{Towards a Multimodal Large Language Model with Pixel-Level Insight for Biomedicine}
\author{
    Xiaoshuang Huang\thanks{Work performed during an internship at Baidu Inc.}\textsuperscript{\rm 1, 2}, Lingdong Shen\textsuperscript{\rm 3}, Jia Liu\textsuperscript{\rm 1}, Fangxin Shang\textsuperscript{\rm 1}, \\
    Hongxiang Li\textsuperscript{\rm 4}, Haifeng Huang\textsuperscript{\rm 1}, Yehui Yang\textsuperscript{\rm 1}\thanks{Corresponding author.}
}
\affiliations{

    \textsuperscript{\rm 1} Baidu Inc \\
    \textsuperscript{\rm 2} China Agricultural University \\
    \textsuperscript{\rm 3} Institute of Automation, Chinese Academy of Sciences \\
    \textsuperscript{\rm 4} Peking University \\
    huangxs497@gmail.com
}

\usepackage{bibentry}

\begin{document}

\maketitle

\begin{abstract}
In recent years, Multimodal Large Language Models (MLLM) have achieved notable advancements, demonstrating the feasibility of developing an intelligent biomedical assistant. However, current biomedical MLLMs predominantly focus on image-level understanding and restrict interactions to textual commands, thus limiting their capability boundaries and the flexibility of usage. In this paper, we introduce a novel end-to-end multimodal large language model for the biomedical domain, named MedPLIB, which possesses pixel-level understanding. Excitingly, it supports visual question answering (VQA), arbitrary pixel-level prompts (points, bounding boxes, and free-form shapes), and pixel-level grounding. We propose a novel Mixture-of-Experts (MoE) multi-stage training strategy, which divides MoE into separate training phases for a visual-language expert model and a pixel-grounding expert model, followed by fine-tuning using MoE. This strategy effectively coordinates multitask learning while maintaining the computational cost at inference equivalent to that of a single expert model. To advance the research of biomedical MLLMs, we introduce the Medical Complex Vision Question Answering Dataset (MeCoVQA), which comprises an array of 8 modalities for complex medical imaging question answering and image region understanding. Experimental results indicate that MedPLIB has achieved state-of-the-art outcomes across multiple medical visual language tasks. More importantly, in zero-shot evaluations for the pixel grounding task, MedPLIB leads the best small and large models by margins of 19.7 and 15.6 respectively on the mDice metric. 
\end{abstract}

\begin{links}
\link{Code}{https://github.com/ShawnHuang497/MedPLIB}
\end{links}

%

\section{Introduction}

Owing to their impressive capabilities in image understanding and text generation, models such as GPT-4V and LLaVA~\cite{liu2024visual} within the realm of Multimodal Large Language Models (MLLMs) have garnered widespread research interest from both academic and industrial sectors~\cite{chen2023minigpt,you2023ferret}. Numerous researchers have dedicated efforts to explore the potential applications of MLLMs in the biomedical field~\cite{wu2023towards,zhang2023pmc,moor2023med}, including LLaVA-Med~\cite{li2024llava} and Med-PaLM M~\cite{tu2024towards}. MLLMs not only generate high-quality responses but also analyze biomedical imagery, demonstrating significant potential to transform traditional medical paradigms~\cite{li2024llava,he2024meddr}. For doctors, such chatbots could significantly alleviate their heavy workloads and enhance efficiency. For patients, it provides more convenient access to professional medical knowledge and advice~\cite{liu2023medical}. Additionally, this could also help alleviate the uneven distribution of medical resources, particularly in regions where they are scarce.

\begin{figure*}[ht]
\centering
\centerline{\includegraphics[width=0.95\textwidth]{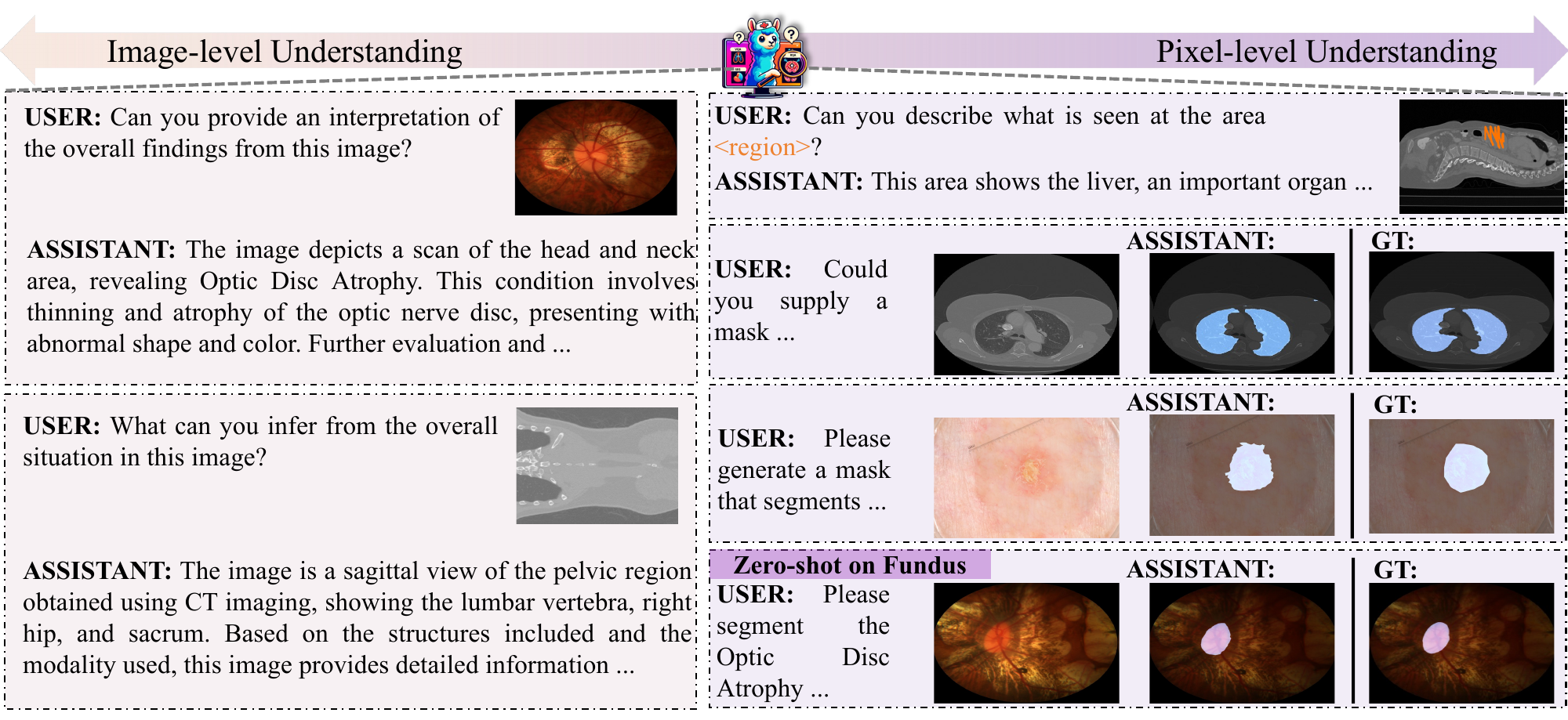}}
\caption{MedPLIB is a biomedical MLLM with a huge breadth of abilities and supports multiple imaging modalities.
} \label{fig:demo}
\end{figure*}

Unlike the image-level VQA of MLLMs in the natural world, the medical domain requires a finer-grained pixel-level understanding to ensure accuracy and answer interpretability. However, existing medical MLLMs~\cite{li2024llava,wu2023towards,moor2023med,tu2024towards} capacity remains restricted to image-level understanding, falling short of pixel-level perceptions.
Pixel-level MLLMs offer several advantages over image-level perception: \textbf{Firstly}, they are capable of recognizing and processing detailed information by focusing on each pixel within an image, such as small lesions or subtle changes in tissue structures. \textbf{Secondly}, they facilitate improved structural recognition and pixel grounding. Pixel-level perception provides more precise grounding outcomes, aiding physicians in better understanding pathological images and formulating treatment plans. \textbf{Thirdly}, pixel-level analysis enables models to understand contextual information on a finer scale. Overall, pixel-level perception is particularly vital in multimodal large models within the medical field, especially in applications requiring high accuracy and detailed recognition. 

Pixel-level MLLMs in biomedicine currently face significant challenges due to these two issues: \textbf{(1) Data scarcity:} Owing to privacy regulations and the high cost of labeling, there is a severe scarcity of pixel-level and complex VQA data. Openly available VQA datasets are typically designed for image-level multiple-choice questions or simple question-and-answer formats, lacking potential for pixel analysis. Meanwhile, segmentation datasets usually contain only segmentation masks and simple category labels, devoid of textual semantic information. \textbf{(2) Models:} Medical VQA often requires a combination of spatial understanding (pixel-level understanding) to ensure confidence and interoperability. Integrating both knowledge-based question-answering and pixel-level analysis within the same MLLM is extremely challenging. Such multimodal input and output not only demand high architectural flexibility from the model but also require the model to balance the knowledge and capabilities of different tasks within a constrained parameter space.

\textbf{For the first challenge}, we propose MeCoVQA dataset. it amasses a substantial collection of segmentation datasets with category labels and, uses a Large Language Model (LLM) in conjunction with manual processing. Specifically, we first convert all segmentation masks of the images into structured metadata, which includes modality, body part, image orientation, and category labels corresponding to mask instances. Then, we use this metadata as a prompt to provide the LLM, which generates a comprehensive description of the image. Finally, we integrate the metadata and image description back as a prompt to the LLM to generate complex question-and-answer data.
\textbf{For the second challenge}, in terms of framework, we expand the LLM's vocabulary and incorporate a region projector to extend the input modalities of the MLLMs. Additionally, inspired by LISA~\cite{lai2024lisa}, we introduce a ``\textless SEG\textgreater'' token to identify and extract features necessary for pixel grounding, combined with SAM-Med2D~\cite{cheng2023sam} to expand the response modalities of the MLLMs. It is noted that the overall framework is end-to-end. To better accommodate tasks of varying granularity, such as pixel grounding tasks and visual question-answering tasks, we introduce a novel multi-granular MoE training strategy within MLLMs that incorporates expert prior knowledge. Specifically, we train two experts separately for VQA and pixel grounding tasks. Subsequently, we integrate the two distinct experts via a training router.

In summary, our contributions are as follows:
\textbf{(1) Model.} We propose an end-to-end MLLM with pixel-level insight for biomedicine, named MedPLIB. It simultaneously supports VQA, pixel-level prompts (points, bounding boxes, and free-form shapes), and pixel-level grounding. Experimental results indicate that MedPLIB achieves state-of-the-art outcomes across multiple medical visual language datasets. 
\textbf{(2) MeCoVQA Dataset.} It comprises an array of 8 modalities with a total of 310k pairs for complex medical imaging question answering and image region understanding.
\textbf{(3) Open-source.} The data, codes, and model checkpoints will be released to the research community.

\begin{figure*}[ht]
\centerline{\includegraphics[width=0.95\textwidth]{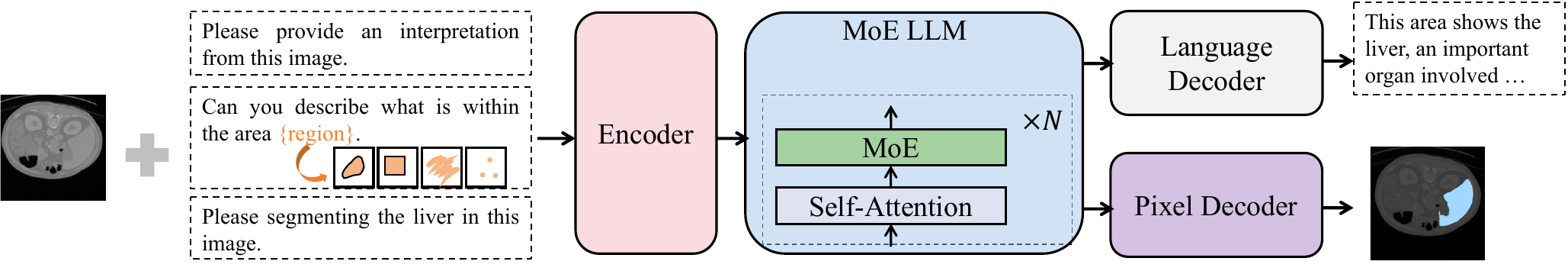}}
\caption{Overview of the proposed MedPLIB. It consists of three parts: encoder, MoE LLM, and decoder.
} \label{fig:model}
\end{figure*}

\section{Related Work}


\newcommand{\cmark}{\textcolor{green!40!black}{\ding{51}}}
\newcommand{\xmark}{\textcolor{red}{\ding{55}}}


\newcommand{\pub}[1]{\color{gray}{\tiny{#1}}}
\newcommand{\gray}[1]{\textcolor{gray}{#1}}

\textbf{Biomedical Visual Question Answering.}
Medical VQA can be categorized into two types based on the scale of the model and data size.
Early approaches employ Convolutional Neural Networks~\cite{simonyan2014very}, long short-term memory models~\cite{hochreiter1997long} followed by feature fusion method~\cite{kim2018bilinear} to generate response. Additionally, the transformer-based approaches like BERT~\cite{devlin2018bert} and BioBert~\cite{lee2020biobert} achieve impressive performance. However, Due to the scarcity of medical data and model size, these models were prone to overfitting, leading to suboptimal robustness~\cite{liu2023cross}. The emergence and success of MLLMs across broad applications have been well explored in the natural world~\cite{liu2024visual,chen2023minigpt,you2023ferret}. Parallel to these developments, the biomedical community has been zealously advancing the capabilities of MLLMs. A focal point of recent research has been the specialized domain of MLLMs, where significant advances have been made, particularly highlighted by models such as RadFM~\cite{wu2023towards}, LLaVA-Med~\cite{li2024llava}, and others~\cite{luo2023biomedgpt,liu2023medical}. These innovations have significantly advanced the biomedical applications of MLLMs.
Although these models show strong task transfer capabilities, they are limited to fixed image and text inputs and outputs. This paper proposes expanding MLLMs to handle diverse input modalities (images, text, free-shape region prompts) and outputs (text, masks) to fully harness their potential.

\textbf{Biomedical Image Segmentation.}
Over the decades, medical image segmentation has evolved significantly, with specialist small models under specific imaging modalities, such as U-Net~\cite{ronneberger2015u}, TransUnet~\cite{chen2021transunet}, and Swin-Unet~\cite{cao2022swin}, achieving commendable results. However, the robustness and generalizability of these specialist models are suboptimal, restricting their application across multiple medical imaging modalities simultaneously. Recent research has shifted focus towards generic medical image segmentation~\cite{zhang2023customized,wu2023medical} and text-guided pixel grounding~\cite{li2023lvit,huang2024cross}, yet both are hindered by data scarcity and model capacity, limiting performance enhancements. Unlike these approaches, our work is pioneering in the medical image analysis domain as we explore the expansion of pixel grounding capabilities within MLLMs, demonstrating the potential to circumvent the limitations of previous methods.

\textbf{Mixture of Experts.}
MoE models have been proposed to augment the number of model parameters without incurring additional computational costs in machine learning~\cite{jacobs1991adaptive,fedus2022switch}. 
A series of approaches naturally decouple experts based on modal categories and pre-define each expert to handle a specific task~\cite{long2023multiway}. This allows for the efficient utilization of shared model parameters. However, it necessitates manual switching between the required experts and lacks adaptive coordination among tasks.
Recently, researchers in the fields of multimodal and natural language processing have focused on the study of soft routers~\cite{chen2024eve}. Soft MoE systems could lead the model to adaptively adjust between experts based on the input data and achieve model sparsity.
The work most relevant to our architecture includes MoE-LLaVA~\cite{lin2024moe} and LLMBind~\cite{zhu2024llmbind}, where all experts possess and are limited to the same prior knowledge. our paper focuses on resolving conflicts among various tasks. We advocate for distinct experts to possess independent task-specific prior knowledge and to coordinate among different tasks effectively.

\begin{figure*}[ht]
\centerline{\includegraphics[width=0.95\textwidth]{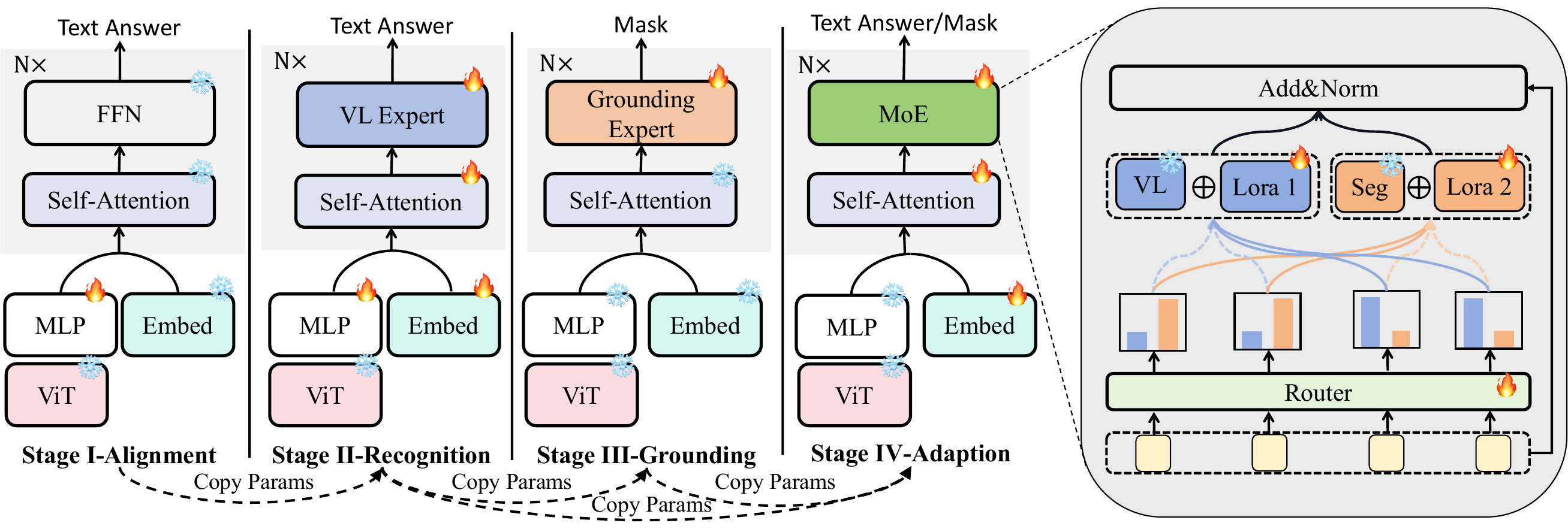}}
\caption{The multi-granular training strategy and the MoE block.
} \label{fig:train}
\end{figure*}

\section{Method}

\subsection{Architecture}
The overall framework of MedPLIB comprises three structural layers: the encoder, the MoE LLM, and the decoder, as illustrated in Figure~\ref{fig:model}. 

\subsubsection{Encoder}
\label{sec:encoder}
The encoder aims to encode all types of inputs (image, text, visual prompt) into a unified feature space. 

\textbf{Vision Tower and Pixel Encoder.} Given the input image $v \in \mathbb{R}^{H\times W \times 3}$, we utilize the pre-trained CLIP visual encoder CLIP-ViT-L/336~\cite{radford2021learning} with a vision-projector as vision tower to extract the feature $V \in \mathbb{R}^{C_v \times N_v}$, where $N_v = \frac{H}{14} \times \frac{W}{14}$ and $C_v$ is the hidden size of vision tower. Then project it as $\hat{V} \in \mathbb{R}^{C_{llm} \times N_v}$, where $C_{llm}$ is the hidden size in the MoE LLM. 
Similarly, we employ a pre-trained Visual Transformer (ViT)~\cite{dosovitskiy2020image} with medical adapter layers~\cite{cheng2023sam} as pixel encoder to get the pixel features $V_p \in \mathbb{R}^{C_p \times N_v}$, where $C_p$ denotes the hidden size in the pixel decoder. 

\textbf{Vision Prompt Encoder.} This block aims to appropriately prompt the LLM with user-specified areas of interest (boxes, points, free shapes) as inputs.  Inspired by SEEM~\cite{zou2024segment}, we define a vision sampler to convert all types of non-textual queries into visual prompts that reside within the same visual embedding space. Assuming the area of interest input is \( R \) and $m$ is the sampling pixel number, the visual prompt features \( V_{vp} \) can be formatted as $V_{vp} = MLP(\phi(V, m))$,
where $\phi$ and $MLP$ is the random sampling function and linear function.

\textbf{Text Prompt Embedding.} Inspire by LLaVA~\cite{liu2024visual}, we expand the tokenizer's vocabulary with ``\textless region\textgreater'' and ``\textless /region\textgreater'' tokens to better integrate visual and textual prompts while distinguishing between their types. Assuming the input text is processed through the text prompt embedding layer to yield textual features $T \in \mathbb{R}^{C_{llm} \times N_t}$, where $N_t$ is the embedded text length. We then use the embeddings of ``\textless region\textgreater'' and ``\textless /region\textgreater'' to encapsulate the visual prompts \( V_{vp} \). These are then embedded into specified positions in the textual feature sequence to obtain $\hat{T} \in \mathbb{R}^{C_{llm} \times \hat{N}_t}$, where $\hat{N}_t = N_t + 1$. 

Finally, we concatenate $\hat{V}$ and $\hat{T}$ to obtain the output $X \in \mathbb{R}^{C_{llm} \times L}$ of the encoding module, where $L = \hat{N}_t + \hat{N}_v$.

\subsubsection{Large Language Model with MoE}
\label{sec:llm}

For the given input \( X \), the operation of the plain feed-forward layer with LoRA can be abstracted as \( \hat{X} = W X + \Delta W X \), where $W \in \mathbb{R}^{C_{llm} \times C_{llm}}$ is the fixed parameter and $\Delta W \in \mathbb{R}^{C_{llm} \times C_{llm}}$ denotes the parameter update in the training phase. 

To better accommodate tasks of varying granularity, such as pixel grounding tasks and visual question-answering tasks, we have introduced the MoE into the Feed Forward Network (FFN) within the LLM. Let's consider the vision-language expert as \( E_{vl} \) and the grounding expert as \( E_{ground} \). The forward process of MoE layer can be formulated as:
\begin{equation}
    \hat{X} = G(X) (\hat{X}_{vl} + \hat{X}_{ground})
\end{equation}
\begin{equation}
    G(X) = Softmax(W_{g}\cdot X)
\end{equation}
\begin{equation}
    \hat{X}_{vl} = E_{vl} X+ \frac{\alpha}{r} \Delta W_0 X
\end{equation}
\begin{equation}
    \hat{X}_{ground} = E_{ground} X + \frac{\alpha}{r} \Delta W_1 X
\end{equation}
where $G(\cdot)$ denotes the router network in the MoE layer and $W_{g}$ is a trainable parameter. The $\alpha$ and $r$ is hyperparameter. The $W_0 = B_0A_0$ and $W_1 = B_1A_1$  where $A_0, A_1 \in \mathbb{R}^{C_{llm} \times r}$, $ B_0, B_1 \in \mathbb{R}^{r \times C_{llm}}$. 
Therefore, each token in \( X \) is processed by the top-1 expert with the highest probability, and the weighted sum is calculated based on the probabilities of the router. 

\subsubsection{Decoder}
\label{sec:decoder}
For text, we use a linear layer as the Decoder, similar to common language models. For pixel-level grounding decoding, we follow LISA~\cite{lai2024lisa}. We extract the last-layer embedding \( \hat{h}_{ground} \) corresponding to the ``\textless SEG\textgreater'' token and utilize the T-projector to obtain \( h_{ground} \). Finally, we use the SAM-Med mask decoder \(\ \gamma \) to obtain the prediction mask $M$. This process can be formulated as $M = \gamma (h_{ground}, V_p)$.

\begin{figure*}[ht]
\centerline{\includegraphics[width=0.95\textwidth]{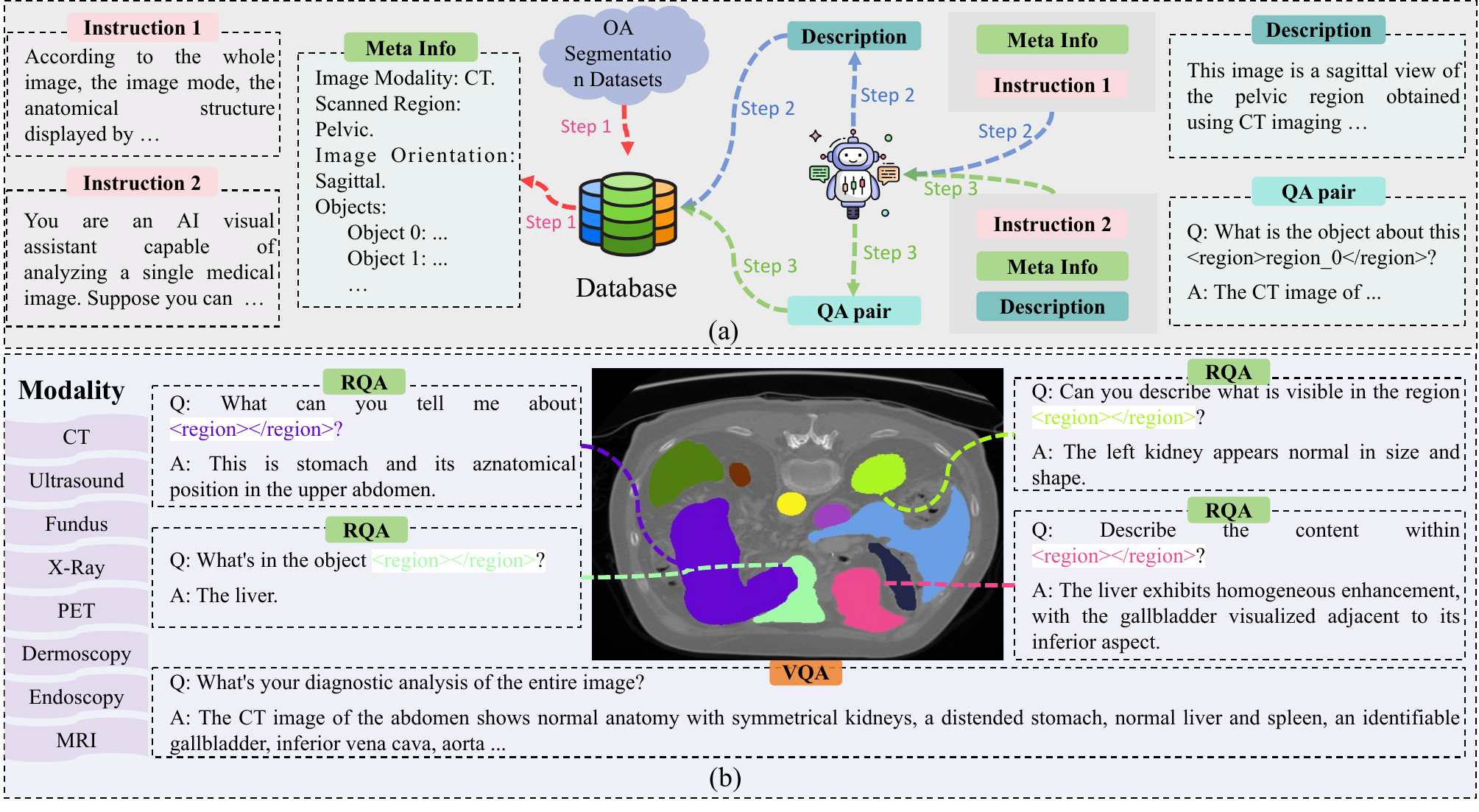}}
\caption{The construction pipeline (a) and a sample (b) of the MoCoVQA dataset.
} \label{fig:data}
\end{figure*}

\subsubsection{Multi-stage Training}

As shown in Figure~\ref{fig:train}, we present the multi-stage training strategy.

Stage I-Alignment:
Following LLaVA-Med~\cite{li2024llava} and LLaVA~\cite{liu2024visual}, we consider only the cross-entropy loss $\mathcal{L}_{reg}$ for text responses during this stage.

Stage II-Recognition:
We train the MLLM as a base model to create an MLLM proficient in medical knowledge and medical imagery understanding. In this stage, we tackle complex visual-language tasks, such as visual knowledge multiple-choice questions, intricate medical Q\&A, and region-based visual question answering. Specifically, we use vast question-and-answer pairs to fine-tune all modules except the vision tower. Thus, we have obtained a MLLM enriched with extensive medical imaging knowledge, where the FFN can be designated as \( E_{vl} \). 
Similar to stage I, the training objective is to minimize the loss $\mathcal{L}_{reg}$.

Stage III-Grounding:
To enhance the pixel grounding capability of the model, we focus on training the grounding expert in this stage. We use the model obtained from stage II as the initial model. We then train using the MeCoVQA-G dataset specifically targeting the FFN layer, pixel decoder, and T-projector. Ultimately, we achieve an MLLM equipped with pixel grounding knowledge, where the FFN is designated as \( E_{ground} \). 
In this stage, we use binary cross-entropy $\mathcal{L}_{bce}$ and dice loss $\mathcal{L}_{dice}$ for pixel grounding losses, and $\mathcal{L}_{reg}$ for the text responses associated with the ``\textless SEG\textgreater'' token.

Stage IV-Adaption:
After completing stage II and stage III, we obtain the parameters for \( E_{vl} \), \( E_{ground} \), and other modules. We then mix all available data and unfreeze all parameters, employing LoRA for fine-tuning through expert mixing. Using the router \( G(\cdot) \), tokens are distributed to different experts for collaborative processing. This approach not only maintains minimal computational expenditure but also preserves the distinct prior knowledge of each expert.
During mixed training, the optimization objective can be formulated as:
\begin{equation}
    \mathcal{L} = \lambda_{reg}\mathcal{L}_{reg} + \lambda_{bce} \mathcal{L}_{bce} + \lambda_{dice} \mathcal{L}_{dice}
\end{equation}
where $\lambda_{reg}$, $\lambda_{bce}$, $\mathcal{L}_{dice}$ are the hyperparameters to balance different objectives.


\begin{table*}[ht]
\small
\centering
\resizebox{0.95\textwidth}{!}{%
\begin{tabular}{c|c|ccccccccccccc}
\toprule

\multicolumn{1}{c|}{\multirow{2}{*}{\textbf{Model}}} & \multicolumn{1}{l|}{\multirow{2}{*}{\textbf{Param.}}} & \multicolumn{9}{c|}{\textbf{OmniMedVQA Benchmark}}                                                                                         & \multicolumn{4}{c}{\textbf{MeCoVQA Test}}                              \\ \cmidrule{3-15} 
\multicolumn{1}{l|}{}        & \multicolumn{1}{l|}{}                          & \textbf{CT} & \textbf{MR} & \textbf{OCT} & \textbf{Der} & \textbf{MIC} & \textbf{X-Ray} & \textbf{FP} & \textbf{US}  & \multicolumn{1}{c|}{\textbf{Mean}} & \multicolumn{2}{c|}{\textbf{MeCoVQA-C}} & \multicolumn{2}{c}{\textbf{MeCoVQA-R}} \\ \midrule
MiniGPT-4~\cite{zhu2023minigpt}     & 7B                                       & 23.67       & 28.65       & 33.62        & 41.28        & 29.31        & 37.15    & 42.46      & 26.42       & 32.82                              & -                & -                & -               & -              \\
BLIP-2~\cite{li2023blip}            & 4B                                     & 59.90       & 43.47       & 69.57        & 40.93        & 51.46        & \textbf{64.97}          &\textbf{67.61} & 39.05       & 54.62                              & -                 & -                 &-                 &-             \\
InstructBLIP~\cite{dai2024instructblip}      & 7B                                     & 29.48       & 36.13       & 45.54        & \textbf{63.02}        & 48.65        & 58.14    & 44.32      & \textbf{43.35}       & 46.08                              &-               &-               &-              &-             \\
LLaVA~\cite{liu2024visual}              & 7B                                    & 18.33       & 28.87       & 37.21        & 49.67        & 28.7        & 28.35  & 34.05        & 23.16       & 31.04                              &30.39                  &41.00                  &18.01                 &44.98                \\
VPGTrans~\cite{zhang2024vpgtrans}       & 7B                                        & 22.88       & 26.47       & 27.23        & 45.42        & 25.53        & 44.18   & 36.83       & 27.49       & 32.00                              &-               &-               &-              &-             \\
RadFM~\cite{wu2023towards}              & 14B                                    & 27.93       & 24.71       & 33.96        & 38.32        & 26.27        & 26.60  & 31.41        & 16.54       & 28.22                              &-               &-               &-              &-             \\
LLaVA-Med~\cite{li2024llava}            & 7B                                  & 19.55       & 30.49       & 38.96        & 46.42        & 29.27        & 32.41     & 43.13     & 30.37       & 33.83                              &19.88                  &33.94                  &15.47                 &34.86                \\
LISA~\cite{lai2024lisa}\textsuperscript{\textdagger}   & 7B                                                &62.96             &49.01             &66.19              &41.61              &54.70              &62.34      &46.71          &32.10             &51.95                                    &56.63                  &52.83                 &12.31              &13.20             \\ \midrule
MedPLIB-w/o MoE                        & 7B               &\textbf{63.22}             & 50.12           &67.24              &43.37              &54.98              &62.87       & 49.55         &33.70             &53.13                     &56.66                  &\textbf{52.94}                  &54.60                 &52.87                \\ 
\textbf{MedPLIB}                   & 12B/7B                    &62.70             &\textbf{66.97}             &\textbf{75.05}              &51.47              &\textbf{64.40}              &60.25        & 65.04        &38.75             &\textbf{60.58}                     &\textbf{58.49}                  &49.41                  &\textbf{64.92}                 &\textbf{63.84}                \\ \bottomrule
\end{tabular}%
}
\caption{Performance on VQA. For closed-set OmniMedVQA~\cite{hu2024omnimedvqa}, we report accuracy metrics. For open-ended MeCoVQA, we report precision (left) and recall (right) metrics. ``CT'', ``MR'', ``OCT'', ``Der'', ``Mic'', ``US'', ``FP'' denote Computed Tomography, Magnetic Resonance Imaging, Optical Coherence Tomography, Dermoscopy, Microscopy Images, Fundus Photography, and Ultrasound, respectively. The "A/B" format in the column of Parameters(Param.) indicates activated parameters during training and inference. \textsuperscript{\textdagger} represents that we implement by office open-source code on our MeCoVQA dataset.}
\label{tab:vqa}
\end{table*}

\begin{table*}[ht]
\small
\centering
\resizebox{0.95\textwidth}{!}{%
\begin{tabular}{c|c|cccc|cccccc}
\toprule
\multicolumn{1}{c|}{\multirow{2}{*}{\textbf{Model}}} & \multicolumn{1}{c|}{\multirow{2}{*}{\textbf{Param.}}} & \multicolumn{4}{c|}{\textbf{MeCoVQA-G Test}}                                                           & \multicolumn{6}{c}{\textbf{Zero-shot}}                                                            \\ \cmidrule{3-12} 
\multicolumn{1}{c|}{}        &\multicolumn{1}{c|}{}                        & \textbf{Der}    & \multicolumn{1}{l}{\textbf{CT}} & \textbf{PET}   & \multicolumn{1}{c|}{\textbf{Mean}} & \textbf{X-Ray} & \textbf{End} & \textbf{MR}    & \textbf{US}    & \textbf{FP}    & \textbf{Mean}  \\ \midrule
LViT~\cite{li2023lvit}~\pub{TMI'23}     & 30M                                            & 84.37          & 58.10                           & 74.45          & 72.31                              & 23.15          & 11.87        & 12.18          & 0.46           & 15.13          & 12.56          \\
ReclMIS~\cite{huang2024cross}~\pub{Arxiv'24}   & 74M/24M                                            & 88.81          & 74.96                           & \textbf{81.15} & 81.64                              & 26.62          & 10.19        & 3.54           & 0.00           & 11.29          & 10.33          \\
LAVT~\cite{yang2022lavt}~\pub{CVPR'22}   & 119M                                              & 92.59          & 77.34                           & 79.13          & 83.02                              & 17.51          & 4.09         & 1.04           & 0.00           & 0.12           & 4.55           \\
DMMI~\cite{hu2023beyond}~\pub{ICCV'23}    & 115                                              & \textbf{93.38} & \textbf{79.97}                  & 80.63          & \textbf{84.66}                     & 18.46          & 0.68         & 1.97           & 0.00           & 0.04           & 4.23           \\
LISA~\cite{lai2024lisa}~\pub{CVPR'24}    & 7B                                              & 81.33          & 52.68                           & 54.20          & 62.74                              & 17.02          & 32.42       & 11.99          & 14.37          & 7.64           & 16.69          \\ \midrule
MedPLIB-w/o MoE   & 7B    & 79.66          & 55.92                           & 56.97          & 64.18                              & 20.92          & 36.43       & 15.87          & 17.98          & 11.31           & 20.50          \\ 
\textbf{MedPLIB}    & 14B/7B                                            & 79.90          & 59.83                           & 64.59          & 68.11                              & \textbf{28.25} & \textbf{44.19}        & \textbf{27.52} & \textbf{35.64} & \textbf{25.76} & \textbf{32.27} \\ \bottomrule
\end{tabular}%
}
\caption{Performance on MeCoVQA-G test set and zero-shot on cross modalities. The MeCoVQA-G test set compromises three modalities (CT, dermoscopy, PET). ``End'' denotes Endoscopy. The "A/B" format in the column of Parameters(Param.) indicates activated parameters during training and inference. All results in this table are implemented by office open-source code on our MeCoVQA-G dataset. }
\label{tab:seg}
\end{table*}

\subsection{MeCoVQA Dataset}
Large models are increasingly used to generate high-quality data, addressing data scarcity. However, in medical imaging, open-source datasets for detailed question-and-answer interactions remain limited. These are vital for intelligent biomedical assistants who need to perform detailed medical analyses and interact with patients. We suggest a new strategy for creating such detailed interactive data.
MoCoVQA was generated through the collaborative efforts of humans and an AI assistant, derived from large-scale biomedical image segmentation datasets. As shown in Figure~\ref{fig:data}, the generation process can be divided into three steps:

\Rmnum{1}. Manually generating instance-level meta information for each image based on its mask. We randomly sampled 100k biomedical images with instance masks from the SA-Med2D-20M~\cite{ye2023sa}. Then we enrich the images with additional details to compile the meta information, which includes modality, scanned region, orientation, and object instances.

\Rmnum{2}. 
We use an AI assistant to get global descriptions for images, adjusting prompts to produce 500 data points per modality, which are manually reviewed for quality. We finalize the prompts only when all points meet quality standards.

\Rmnum{3}. Utilizing the AI assistant to craft pixel-level conversations based on the meta information and global descriptions obtained in step \Rmnum{2}. At this step, we used complex instructions to generate diverse data, manually refining prompts multiple times, as in stage II, to ensure quality.

The MeCoVQA dataset could be divided into three subsets: MeCoVQA-C (MeCoVQA-Complex), MeCoVQA-R (MeCoVQA-Region), and MeCoVQA-G (MeCoVQA-Grounding), which are used for complex VQA, region VQA, and pixel grounding, respectively. Complex VQA and region VQA are constructed through the aforementioned pipeline. MeCoVQA-G is generated by specifying question templates combined with mask category labels. Overall, the training set numbers for MeCoVQA-C, MeCoVQA-R, and MeCoVQA-G are 80k, 126k, and 100k respectively.
Additionally, the numbers of their corresponding test sets are 1477, 2633, and 2344, respectively. For more information about MeCoVQA datasets, please refer to the Appendix.


\begin{table*}[ht]
\small
\centering
\begin{tabular}{c|cccc|cccc|c}
\toprule
&\textbf{stage I} & \textbf{stage II} & \textbf{stage III} & \textbf{stage IV} & \textbf{MeCoVQA-C} & \textbf{MeCoVQA-R} & \textbf{OmniMedVQA} & \textbf{MeCoVQA-G} & \textbf{Mean}  \\ \midrule
(a)&    -            &-                 &-                 &-                 & 60.18              & 58.88              & 53.30               & 34.11                 & 51.62          \\
(b)&               &                 &                 & \cmark          & 58.01              & 62.91              & 57.55               & 38.48                 & 54.24          \\
(c)&\cmark          &                 &                 & \cmark          & {58.76}     & \textbf{63.74}     & \textbf{57.95}      & 37.22                 & 54.42          \\
(d)&\cmark          & \cmark          &                 & \cmark          & 58.25              & 63.24              & 57.00               & 39.63                 & 54.53          \\
(e)&\cmark          &                 & \cmark          & \cmark          & 58.20              & 57.91              & 56.00               & \textbf{47.52}        & 54.91          \\
(f)&\cmark          & \cmark          & \cmark          & \cmark          & \textbf{60.55}     & 62.14              & 57.90                & 46.73        & \textbf{56.83} \\ \bottomrule
\end{tabular}%
\caption{Effect of the different training stages.}
\label{tab:stage}
\end{table*}

\makeatletter
  \newcommand\figcaption{\def\@captype{figure}\caption}
  \newcommand\tabcaption{\def\@captype{table}\caption}
\makeatother

\begin{figure*}[ht]
\begin{minipage}[b]{0.47\textwidth}
\small
\centering
\resizebox{\textwidth}{!}{%
\begin{tabular}{cc|ccccc}
\toprule
\textbf{CF} & \textbf{Top-k} & \textbf{\begin{tabular}[c]{@{}c@{}}MeCo-\\ VQA-C\end{tabular}} & \textbf{\begin{tabular}[c]{@{}c@{}}MeCo-\\ VQA-R\end{tabular}} & \textbf{\begin{tabular}[c]{@{}c@{}}OmniM-\\ edVQA\end{tabular}} & \textbf{\begin{tabular}[c]{@{}c@{}}MeCo-\\ VQA-S\end{tabular}} & \textbf{Mean}  \\ \midrule
1           & 1              & 57.95                                                         & 59.23                                                         & 55.85                                                           & 44.92                                                             & 54.49          \\
1.5         & 1              & \textbf{60.55}                                                & \textbf{62.14}                                                & \textbf{57.9}                                                   & \textbf{46.73}                                                    & \textbf{56.83} \\
2           & 1              & 55.35                                                         & 55.94                                                         & 56.15                                                           & 46.64                                                             & 53.52          \\
2           & 2              & 55.05                                                         & 57.64                                                         & 54.35                                                           & 45.32                                                             & 53.09          \\ \midrule
\end{tabular}%
}
\tabcaption{Effect of the hyper-parameters. ``CF'' denotes the capability factor of the expert. ``Top-k'' represents distributing the token to the top k experts with the highest probabilities for processing.}
\label{tab:hyperparam}
\end{minipage}
\hfill
\begin{minipage}[b]{0.47\textwidth}
    \centering
    \includegraphics[width=1.\textwidth]{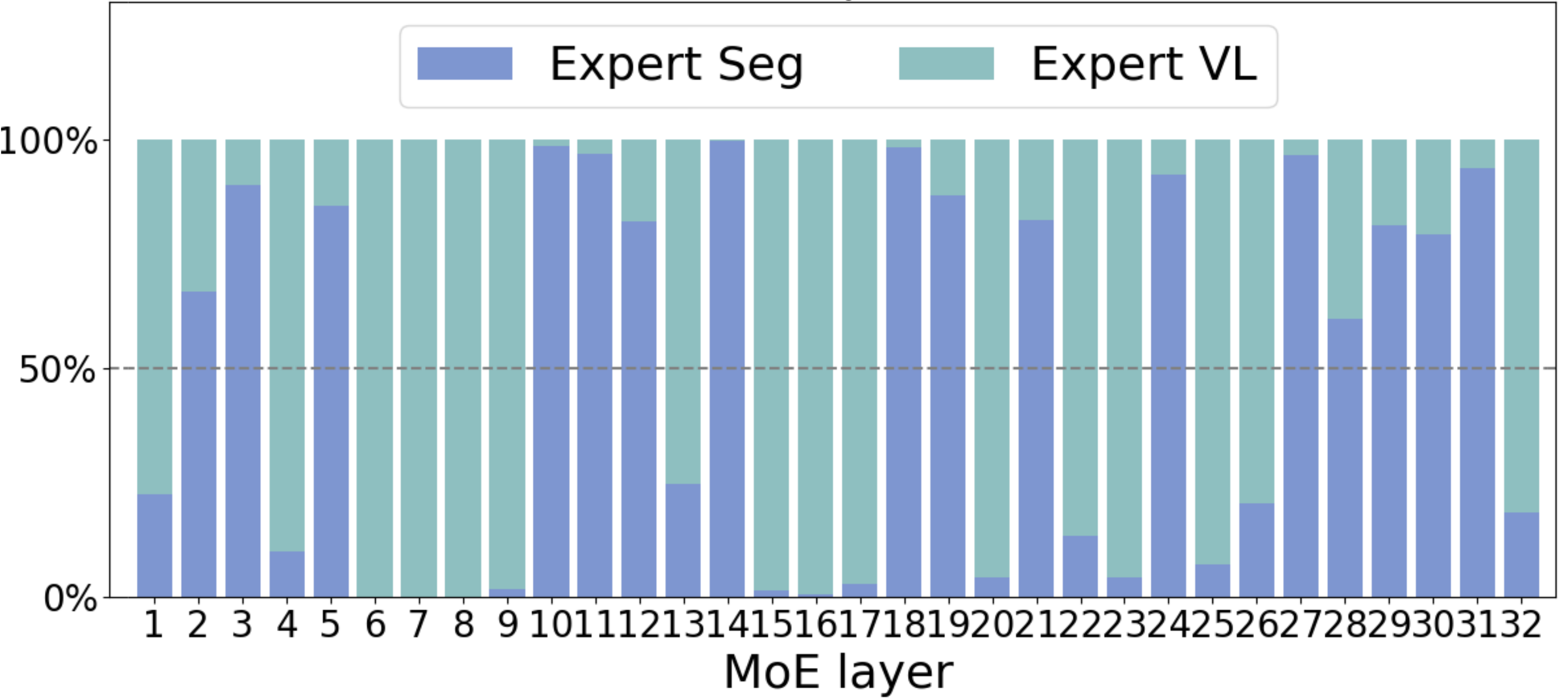}
    \caption{Distribution of tokens among different experts.}\label{ablation: mask_ratio}
\label{fig:ep_vis}
 \end{minipage}
\end{figure*}

\section{Experiments}
\label{sec:exp}
\subsection{Experimental Setup}
\textbf{Model Settings.}
We employ SAM-Med2D~\cite{cheng2023sam} as the pixel encoder and mask decoder. We use LLaMA-7B~\cite{touvron2023llama} as a base LLM. Following LLaVA 1.5~\cite{liu2024visual}, we utilize CLIP-Large~\cite{radford2021learning} as the vision tower and the MLP consists of two linear layers with GELU activation function~\cite{hendrycks2016gaussian}. The parameters of the model with 2 experts are 12 Billion. The training durations for stages I to IV are 9, 17, 15, and 77 hours, respectively. We provide additional model details in the Appendix.

\textbf{Datasets.}
For the training data in stage I, we employ LLaVA-Med-alignment~\cite{li2024llava}. We utilize the union of MeCoVQA-R, MeCoVQA-C, SLAKE~\cite{liu2021slake}, PathVQA~\cite{he2021towards}, PMC-VQA~\cite{zhang2023pmc}, ImageClef2021~\cite{ben2021overview}, ImageClef2019~\cite{abacha2019vqa}, and VQA-RAD~\cite{lau2018dataset} in stage II. In stage III, we use MeCoVQA-G for training. Finally, we employ all data used in stages II and III for training in stage IV. The data volumes for stages I to IV are 330k, 400k, 100k, and 500k, respectively.

\textbf{Metrics.}
For closed-set VQA, we report accuracy. For open-set VQA, we report precision and recall. For pixel-level grounding, we report the mean dice score.

\subsection{Performance Evaluation}

\textbf{Performance on VQA Benchmark.} OmniMedVQA~\cite{hu2024omnimedvqa} is a large medical VQA benchmark that utilizes single-choice questions. We present the evaluation results of the open-source portion of the OmniMedVQA benchmark in Table~\ref{tab:vqa}. Across seven modalities, MedPLIB leads the second-best model, BLIP-2~\cite{li2023blip}, by an advantage of 7.84 points in the mean performance. Additionally, our MedPLIB significantly outperforms other biomedical MLLMs. For more analysis of this table, please refer to the Appendix.


\textbf{Complex VQA Evaluation.}
Compared to test sets like OmniMedVQA~\cite{hu2024omnimedvqa}, MeCoVQA-C features longer open-ended questions. As indicated in the third and fourth columns at the end of Table \ref{tab:vqa}, our MedPLIB achieves better precision compared to LISA~\cite{lai2024lisa}, but slightly lower recall. We believe this is due to MedPLIB balancing its pixel grounding capabilities, which slightly compromises its ability to handle long-text VQA tasks.

\textbf{Region-level VQA Evaluation.}
Region-level VQA demands pixel-level image understanding. Current models lack support for region-specific prompts. To enable this, we integrate coordinates into the prompts for models like LLaVA~\cite{liu2024visual}, LLaVA-Med~\cite{li2024llava}, and LISA~\cite{lai2024lisa}. As demonstrated in the last two columns of Table \ref{tab:vqa}, our MedPLIB significantly outperforms these models.

\textbf{Pixel Grounding Evaluation.}
Since there has not yet been a biomedical MLLM with pixel grounding capabilities, we compare our MedPLIB with small models that possess pixel grounding capabilities and with the influential LISA~\cite{lai2024lisa} from the general domain. As shown in the second column of Table~\ref{tab:seg}, our MedPLIB surpasses LISA~\cite{lai2024lisa} by 5.37 points on the mDice metric in the MeCoVQA-G. However, it performs closely to the small model LVIT~\cite{li2023lvit} and significantly lags behind DMMI~\cite{hu2023beyond}.

\textbf{Zero-shot to Pixel Grounding.}
To evaluate the generalization capabilities of our model, we conducted zero-shot assessments on five medical imaging modalities that the model did not see. As shown in the last six columns of Table~\ref{tab:seg}, our MedPLIB demonstrated remarkable generalization capabilities. It significantly outperformed the best model on the MeCoVQA-G Test set, LViT~\cite{li2023lvit}, and pixel-grounding MLLM (LISA~\cite{lai2024lisa}). This underscores the substantial potential of our approach in addressing the generalization challenges that small models for medical image grounding struggle to overcome.

\subsection{Qualitative Results}
Figure~\ref{fig:demo} illustrates the performance of MedPLIB across various capabilities, addressing many issues beyond the scope of existing biomedical MLLMs. Additionally, we visualized the distribution of tokens among different experts in Figure~\ref{fig:ep_vis}. Overall, each expert processed about half of the tokens. This indicates that in MedPLIB, \(E_{vl}\) and \(E_{ground}\) achieved a good level of collaboration and load balancing. Additionally, we provide more results in the Appendix.

\subsection{Ablation Study}
We investigated the impact of MoE, training stages, and key hyperparameters on model performance in this section. It is important to note that conducting ablation experiments on all data is prohibitively costly, so during the ablation stage, we used a training set consisting of 20k samples extracted from the total dataset. For testing, we extracted 400 samples from the original MeCoVQA-C and MeCoVQA-R test sets and 2000 samples from OmniMedVQA~\cite{hu2024omnimedvqa}.

\textbf{Effect of MoE.}
\label{sec:effect_of_moe}
The variants (a) and (b) in Table \ref{tab:stage} display the performance of using a standard FFN and a MoE, respectively. Overall, the average performance of MoE across four datasets is 2.62 points higher than that of FFN, demonstrating MoE's adaptability to our tasks.

\textbf{Effect of Multi-stage Training.}
\label{sec:effect_of_stage}
In Table \ref{tab:stage}, we conduct five variant experiments to demonstrate the rationale of our multi-stage tuning. Following LLaVA-Med~\cite{li2024llava}, we samely use stage I to align visual features to text embedding space. Variants (b) and (c) indicate that having alignment is more beneficial for fine-tuning in stage IV. Variants (c) and (d) demonstrate that using the $E_{vl}$ from stage II as the initial weights for stage IV helps the model focus more on VL tasks. Similarly, variants (c) and (e) show that using the $E_{ground}$ from stage III as the initial weights for stage IV help the model focus more on VL tasks. Lastly, the tuning strategy in stage IV as per variant (f), compared to variant (b), allows the model to better balance image-level and pixel-level tasks.

\textbf{Effect of Top-k.}
We explored the impact of using top-1 and top-2 routing on model performance. Utilizing top-2 in our experiments implies equivalence to a dense model (where each expert processes all tokens) since we are using only two experts. The last two rows of Table \ref{tab:hyperparam} indicate that a dense model is less effective than a sparse activated model.

\textbf{Effect of Capability Factor.}
We examined the impact of the Capacity Factor (CF). At CF=1, each expert handles up to half the tokens, risking information loss due to their prior knowledge. At CF=2, experts can process all tokens, leading to noise and redundancy. Empirical evidence suggests CF=1.5 is optimal, balancing the reduction of information loss and noise in token distribution.


\section{Conclusion}
\label{sec:conclusion}
In this paper, we present MedPLIB, a multimodal large language model with pixel-level insight for biomedicine. MedPLIB features flexible inputs and outputs, thereby supporting multiple tasks and creating a more versatile and patient-friendly MLLM. 
To achieve the mentioned targets, we have made efforts on both the model and data levels. On the model level, we introduce a three-layer architecture and a novel MoE training strategy within MLLMs that incorporates expert prior knowledge. On the data level, we introduce the MeCoVQA, which comprises an array of 8 modalities for answering complex medical imaging questions, understanding image regions, and pixel grounding. Experimental results indicate that MedPLIB has achieved state-of-the-art outcomes on the OmniMedVQA benchmark and MeCoVQA test sets. Moreover, MedPLIB has demonstrated encouraging performance in its zero-shot ability for pixel-level grounding.


\newpage
\begin{appendices}
    \section{Appendix}  

    \section{Model Architecture}
\label{sec:model_architecture}
\begin{figure*}[h]
\centerline{\includegraphics[width=1.0\textwidth]{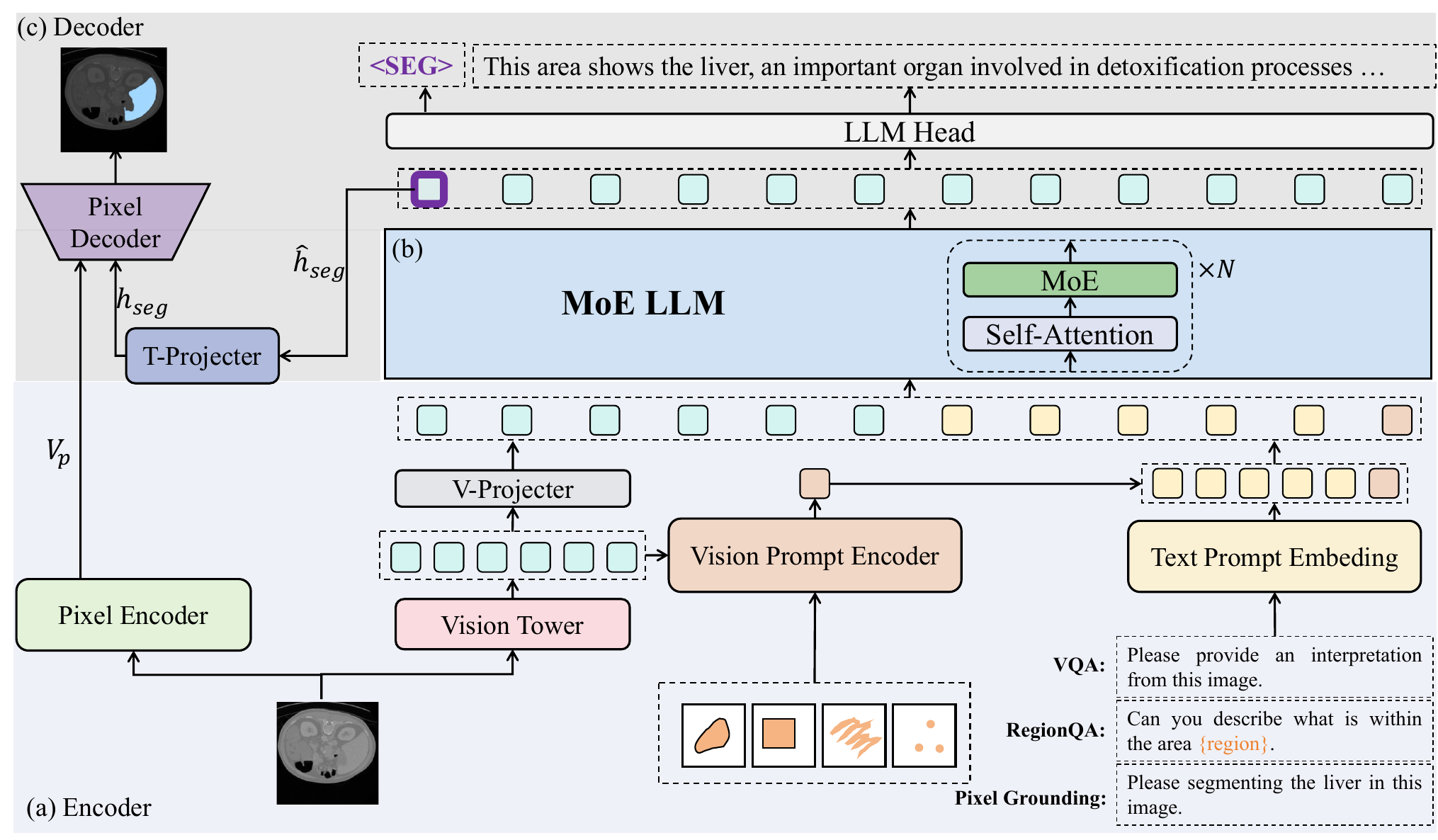}}
\caption{The detailed architecture of the MedPLIB model.
} \label{fig:detail_model}
\end{figure*}

In Figure~\ref{fig:detail_model}, we present a more detailed diagram of the MedPLIB model architecture.

\section{Model Setting Details}
\label{sec:model_set_details}
This project was conducted on Pytorch. We employed SAM-Med2D~\cite{cheng2023sam} as the pixel encoder and mask decoder. The weight of the text generation loss $\lambda_{reg}$ is set to $1.0$, and those of the bce loss $\lambda_{bce}$ and the dice loss $\lambda_{dice}$ are set to $2.0$ and $0.5$, respectively. Following LLaVA 1.5~\cite{liu2024visual}, we utilize CLIP-Large~\cite{radford2021learning} as the vision tower and the MLP consists of two linear layers with GELU activation function~\cite{hendrycks2016gaussian}. We adopt 4 NVIDIA 40G A100 GPUs for training and the training scripts are based on the deepspeed engine. Moreover, the batch size per device is set to 4, and the gradient accumulation step is set to 8. The fine-tuning epochs of stages I to IV are 1, 3, 10, and 3, respectively. We employ ChatGPT-3.5 as the AI assistant.

\section{Model Capability Comparison}
\label{sec:model_capa_compare}
As shown in Table~\ref{tab:capa}, MedPLIB is a biomedical MLLM with a huge breadth of abilities and supports multiple imaging modalities. Not only can it perform image-level visual language tasks like VQA, but it also facilitates question-answering at the pixel level.

Pixel-level understanding has two main aspects:

1. Users can specify image pixels in a free shape to pose questions (i.e., the RQA task). The last two rows in Table 1 show that our method significantly outperforms the baseline and other approaches.

2. The model has pixel-level grounding capability. The small models (e.g., LViT~\cite{li2023lvit}, ReclMIS~\cite{huang2024cross}, LAVT~\cite{yang2022lavt}, DMMI~\cite{hu2023beyond}) are specialized small models designed to perform single tasks, while the large models (e.g., LISA~\cite{lai2024lisa}, Baseline, MedPLIB) are MLLMs. It is clear that all MLLMs perform below the specialists because they must also handle other challenging tasks (e.g., VQA, RQA), whereas specialists focus solely on their assigned tasks. This phenomenon is common across all-in-one generalist models. Moreover, our generalist model demonstrates remarkable zero-shot generalization capability in this task, which is especially valuable in complex real-world applications.

\begin{table*}[h]
\centering
\begin{tabular}{cccccccc}
\toprule
\multicolumn{1}{c|}{\multirow{2}{*}{Method}} & \multicolumn{1}{c|}{\multirow{2}{*}{Image}} & \multicolumn{4}{c|}{Referring Format}                      & \multicolumn{2}{c}{Response Format} \\ \cmidrule{3-8} 
\multicolumn{1}{c|}{}                        & \multicolumn{1}{c|}{}                       & Text   & Box    & Sketch & \multicolumn{1}{c|}{Free shape} & Mask             & Text            \\ \midrule
LViT~\cite{li2023lvit}~\pub{TMI'2023}                                         & \cmark                                      & \cmark & \xmark & \xmark & \xmark                          & \cmark           & \xmark          \\
SAM-Med2D~\cite{cheng2023sam}~\pub{Arxiv'2023}                                    & \cmark                                      & \xmark & \cmark & \cmark & \xmark                          & \cmark           & \xmark          \\
LISA~\cite{lai2024lisa}~\pub{CVPR'2024}                                         & \cmark                                      & \cmark & \xmark & \xmark & \xmark                          & \cmark           & \cmark          \\
RadFM~\cite{wu2023towards}~\pub{Arxiv'2023}                                        & \cmark                                      & \cmark & \xmark & \xmark & \xmark                          & \xmark           & \cmark          \\
MediTron~\cite{li2023lvit}~\pub{Arxiv'2023}                                     & \xmark                                      & \cmark & \xmark & \xmark & \xmark                          & \xmark           & \xmark          \\
Med-PaLM M~\cite{tu2024towards}~\pub{Arxiv'2023}                                   & \cmark                                      & \cmark & \xmark & \xmark & \xmark                          & \xmark           & \xmark          \\
LLaVA-Med~\cite{li2024llava}~\pub{NeurIPS'2023}                                    & \cmark                                      & \cmark & \xmark & \xmark & \xmark                          & \xmark           & \cmark          \\ \midrule
\textbf{MedPLIB}                           & \cmark                                      & \cmark & \cmark & \cmark & \cmark                          & \cmark           & \cmark          \\ \toprule
\end{tabular}
\caption{Comparative analysis of the capabilities across different models. Not only can MedPLIB perform image-level visual language tasks like VQA, but it also facilitates question answering at the pixel level. MedPLIB features flexible inputs and outputs, thereby supporting multiple tasks and creating a more versatile and patient-friendly MLLM.}
\label{tab:capa}
\end{table*}

\section{Comparison of MeCoVQA and Other Datasets}
\label{sec:compara_mecovqa_with_other}

Table~\ref{tab:compare_dataset} presents a comparison between our MeCoVQA dataset and existing datasets. It is evident that MeCoVQA surpasses current methods in terms of the number of modalities, granularity, and the quantity of questions and answers.
\begin{table*}[h]
\centering
\resizebox{\textwidth}{!}{%
\begin{tabular}{ccccccccc}
\toprule
\textbf{Dataset}                              & \textbf{Image Level} & \textbf{Pixel Level} & \textbf{Modalities}      & \textbf{Train} & \textbf{Val} & \textbf{Test} & \textbf{QA-pairs} & \textbf{Images} \\ \midrule
VQA-RAD~\cite{lau2018dataset}                                       & \cmark               & \xmark               & 3                        & 3k             & -            & 0.5k          & 3.5k              & 0.3k            \\
SLAKE~\cite{liu2021slake}                                         & \cmark               & \xmark               & {\color[HTML]{4D4D4D} 3} & 11k            & 1.5k         & 1.5k          & 14k               & 0.6k            \\
PathVQA~\cite{he2021towards}                                       & \cmark               & \xmark               & 2                        & 19k            & 9.5k         & 6k            & 33k               & 5k              \\
PMC-VQA~\cite{zhang2023pmc}                                       & \cmark               & \xmark               & -                        & 153k           & -            & 33k           & 186k              & 164k            \\
{\color[HTML]{191B1F} ImageClef-2021-VQA-Med~\cite{ben2021overview}} & \cmark               & \xmark               & 5                        & 5k             & 0.5k         & 0.5k          & 6k                & 5k              \\
{\color[HTML]{191B1F} ImageClef-2019-VQA-Med~\cite{abacha2019vqa}} & \cmark               & \xmark               & 5                        & 13k            & 2k           & 0.5k          & 15k               & 4k              \\ \midrule
\textbf{MeCoVQA}                              & \cmark               & \cmark               & \textbf{8}               & \textbf{305k}  & -            & \textbf{6k}   & \textbf{311k}     & \textbf{177k}   \\ \bottomrule
\end{tabular}%
}
\caption{A comparison of different open-source medical VQA datasets: "Image level" implies that it includes global image question and answer types, while "Pixel level" means it encompasses region-specific question and answer types as well as pixel grounding types.}
\label{tab:compare_dataset}
\end{table*}

\section{More Information about MeCoVQA Datasets}
\label{sec:more_info_mecovqa}
\textbf{More construction details in Step II of Data Construction.}
Tactics for data quality: 
a. Two experts manually edit 4 templates per modality based on the provided information, then supply these high-quality templates to ChatGPT as prompts to ensure responses align with expert specifications.
b. All three QA types require approval of 1,600 data points (200 per modality) by 2 experts, with iterative improvements to the instruction based on expert feedback until all requirements are met.

Quality standards: a. The description must include all organs or disease categories present in the image with 100\% accuracy, as organized in Step I. b. Randomly sample 4,000 data points (500 per modality) for approval by two experts.

\textbf{More construction details in step III of Data Construction.}
We argue that quality control is a critical aspect of AI-assisted dataset generation. Below is a brief outline of the quality strategies and standards for Step III of the construction pipeline.

\textbf{Strategies:}
\begin{itemize}
    \item Coarse-to-fine: Begin by generating objective global descriptions in Step II, followed by simulating dialogue in Step III.
    \item Templates: Two experts manually edit four templates per modality per task based on the provided information, then supply these high-quality templates to ChatGPT as prompts to ensure responses align with expert specifications.
    \item Manual intervention: Both types require the approval of 3200 data points (200 per modality per task) by two experts, with iterative improvements to the instructions based on expert feedback until all requirements are met.
\end{itemize}

\textbf{Standards:}
\begin{itemize}
    \item For RQA: Since the questions are related to organ or disease categories (e.g., "What's in the object <region>?"), the answer must contain the corresponding category or disease name with 100\% accuracy.
    \item For complex VQA: This involves descriptions of the overall image context, and all organs or disease categories within the image must be included in the answer with 100\% accuracy.
    \item For both types (RQA and complex VQA): The sampled 8,000 data points (500 per modality per task) must be approved by two experts.
    \item For MeCoVQA-G, Experts create two grounding templates, each with two questions and instructions (e.g., "Please segment the {class} in this image"). ChatGPT generates 10 questions and instructions based on these templates, reviewed by experts. The corresponding mask categories are inserted into randomly selected questions with answers matching the mask annotations.
\end{itemize}

\textbf{Quantitative analysis.}
In Table~\ref{tab:mecovqa_img_nums} and Table~\ref{tab:mecovqa_pairs}, we provide detailed data on the number of images and question-answer pairs for different modalities within MeCoVQA, respectively.
\begin{table*}[h]
\centering
\resizebox{\textwidth}{!}{%
\begin{tabular}{cccccccccc}
\toprule
\textbf{Name of Subset}                   & \textbf{CT} & \textbf{Dermoscopy} & \textbf{Endoscopy}         & \textbf{Fundus} & \textbf{MRI} & \textbf{PET}   & \textbf{Ultrasound} & \multicolumn{1}{c}{\textbf{X-Ray}} & \textbf{Total} \\ \midrule
MeCoVQA-C Training                        & 16699       & 6699                & 4073                       & 1061            & 24444        & 5248           & 2643                & 5028                                & 65895          \\
MeCoVQA-C Test                            & 200         & 200                 & {\color[HTML]{4D4D4D} 200} & 83              & 200          & 200            & 194                 & 200                                 & 1477           \\
MeCoVQA-R Training                        & 18470       & 7021                & 3906                       & 1061            & 24332        & 5196           & 2766                & 4646                                & 67398          \\
MeCoVQA-R Test                            & 500         & 500                 & 295                        & 83              & 500          & 57             & 202                 & 496                                 & 2633           \\
{\color[HTML]{191B1F} MeCoVQA-G Training} & 25147       & 7028                & -                          & -               & -            & 5249           & -                   & -                                   & 37424          \\
{\color[HTML]{191B1F} MeCoVQA-G Test}     & 577         & 561                 & 24                          & 800               & 313            & 783            & 169                   & 654                                   & 1921           \\
Total                                     & 61593       & 22009               & 8474              & 2288   & 49476        & 16733 & 5805      & 10370                     & 176748         \\ \bottomrule
\end{tabular}%
}
\caption{Quantification of image counts across different modalities in various subsets of MeCoVQA.}
\label{tab:mecovqa_img_nums}
\end{table*}
\begin{table*}[h]
\centering
\resizebox{\textwidth}{!}{%
\begin{tabular}{cccccccccc}
\toprule
\textbf{Name of Subset}                   & \textbf{CT} & \textbf{Dermoscopy} & \textbf{Endoscopy}         & \textbf{Fundus} & \textbf{MRI} & \textbf{PET}   & \textbf{Ultrasound} & \multicolumn{1}{c}{\textbf{X-Ray}} & \textbf{Total} \\ \midrule
MeCoVQA-C Training                        & 18512       & 6742                & 5081                       & 1068            & 27378        & 9970           & 2644                & 7562                                & 78957          \\
MeCoVQA-C Test                            & 200         & 200                 & {\color[HTML]{4D4D4D} 200} & 83              & 200          & 200            & 194                 & 200                                 & 1277           \\
MeCoVQA-R Training                        & 36090       & 14000               & 7087                       & 2122            & 46823        & 5875           & 5532                & 7947                                & 125476         \\
MeCoVQA-R Test                            & 500         & 500                 & 295                        & 83              & 500          & 57             & 202                 & 500                                 & 2637           \\
{\color[HTML]{191B1F} MeCoVQA-G Training} & 88205       & 7028                & -                          & -               & -            & 5249           & -                   & -                                   & 100482         \\
{\color[HTML]{191B1F} MeCoVQA-G Test}     & 1000        & 561                 & -                          & -               & -            & 783            & -                   & -                                   & 2344           \\
Total                                     & 144507      & 29031               & 12663             & 3356   & 74901        & 22134 & 8572       & 16205                    & 311169         \\ \bottomrule
\end{tabular}%
}
\caption{Quantitative analysis of multimodal question-answering pairs across different modalities in MeCoVQA.}
\label{tab:mecovqa_pairs}
\end{table*}

\section{More Analysis about Table 1 in the Main Paper}
The training data contains relatively low quantities of Der, X-ray, and US modalities, creating an imbalanced dataset that impacts model performance. This underscores the need for more VQA data in these modalities, as noted in OmniMedVQA~\cite{hu2024omnimedvqa}.

For modalities like Dermoscopy, with distributions similar to general domain images, the superior performance of BLIP-2~\cite{li2023blip} and InstructBLIP~\cite{dai2024instructblip} in certain modalities is due to their training on datasets of tens of millions, giving them stronger generalization capabilities.

Much of the data used by BLIP-2~\cite{li2023blip} and InstructBLIP~\cite{dai2024instructblip} was sourced via web scraping, which includes some medical images, especially common modalities like Dermoscopy.

\section{More Qualitative Results}
\label{sec:more_quat_res}
\begin{figure*}[h]
\centerline{\includegraphics[width=1.0\textwidth]{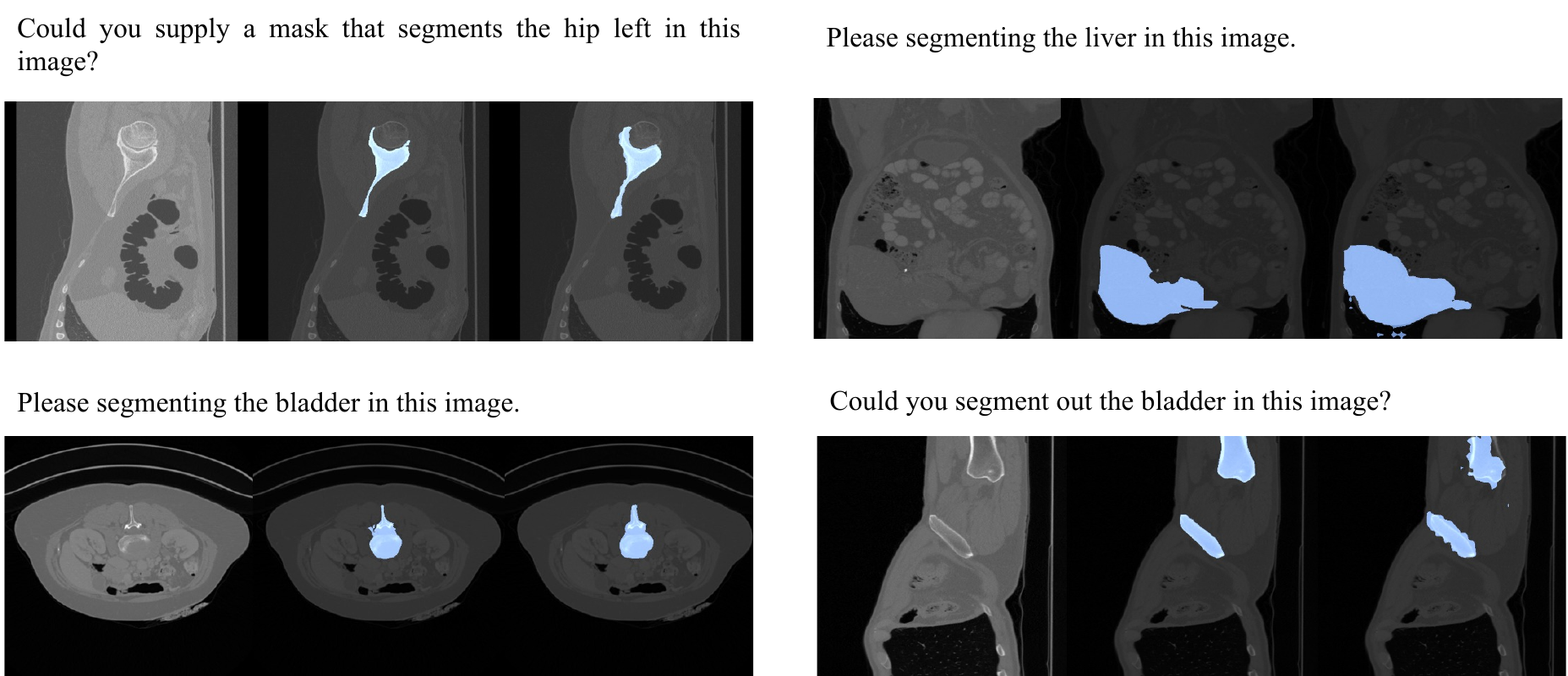}}
\caption{The demonstration of MedPLIB's grounding effects on the CT modality within MeCoVQA-G. The input text prompt is located at the top, with the original input image on the left, the grounding ground truth in the middle, and MedPLIB's grounding results on the right.
} \label{fig:vis_ct}
\end{figure*}

\begin{figure*}[h]
\centerline{\includegraphics[width=1.0\textwidth]{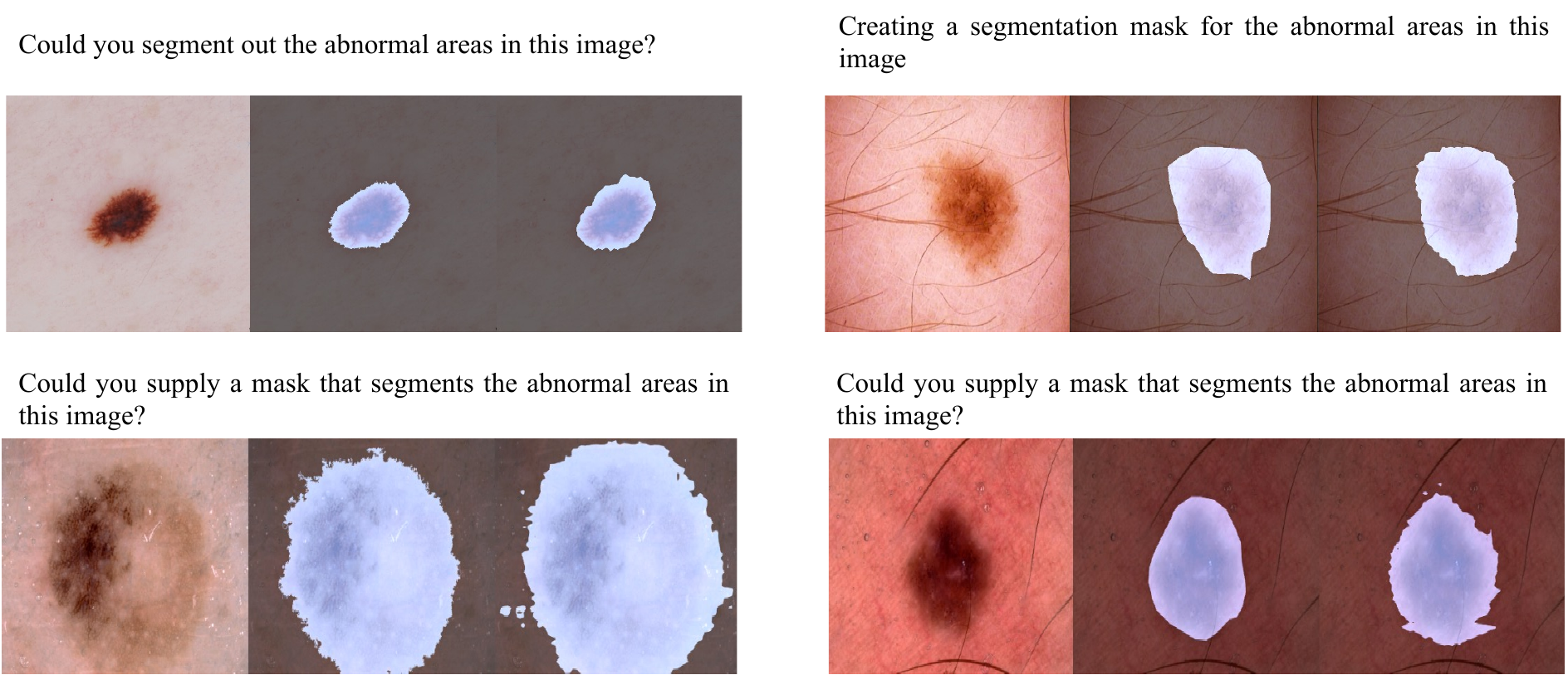}}
\caption{The demonstration of MedPLIB's grounding effects on the dermoscopy modality within MeCoVQA-G. The input text prompt is located at the top, with the original input image on the left, the grounding ground truth in the middle, and MedPLIB's grounding results on the right.
} \label{fig:vis_der}
\end{figure*}

\begin{figure*}[h]
\centerline{\includegraphics[width=1.0\textwidth]{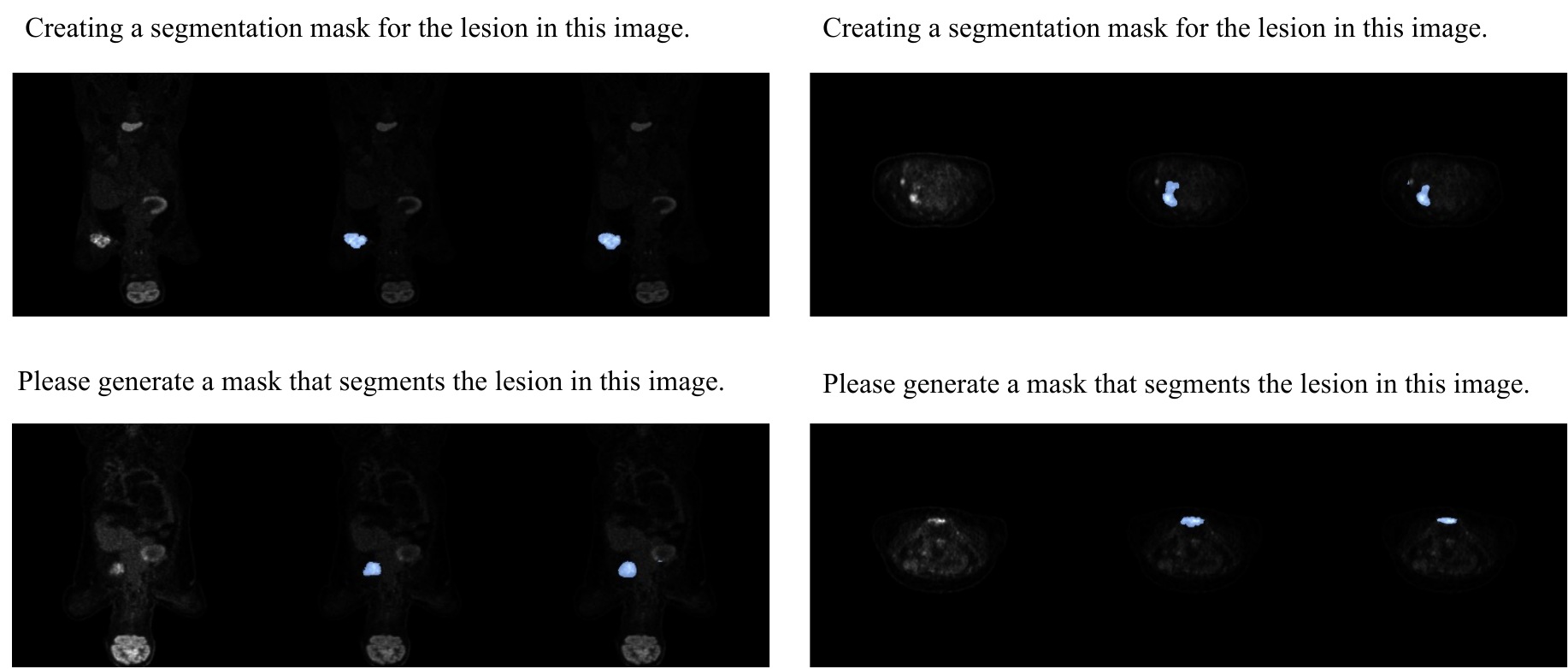}}
\caption{The demonstration of MedPLIB's grounding effects on the PET modality within MeCoVQA-G. The input text prompt is located at the top, with the original input image on the left, the grounding ground truth in the middle, and MedPLIB's grounding results on the right.
} \label{fig:vis_pet}
\end{figure*}

\begin{figure*}[h]
\centerline{\includegraphics[width=1.0\textwidth]{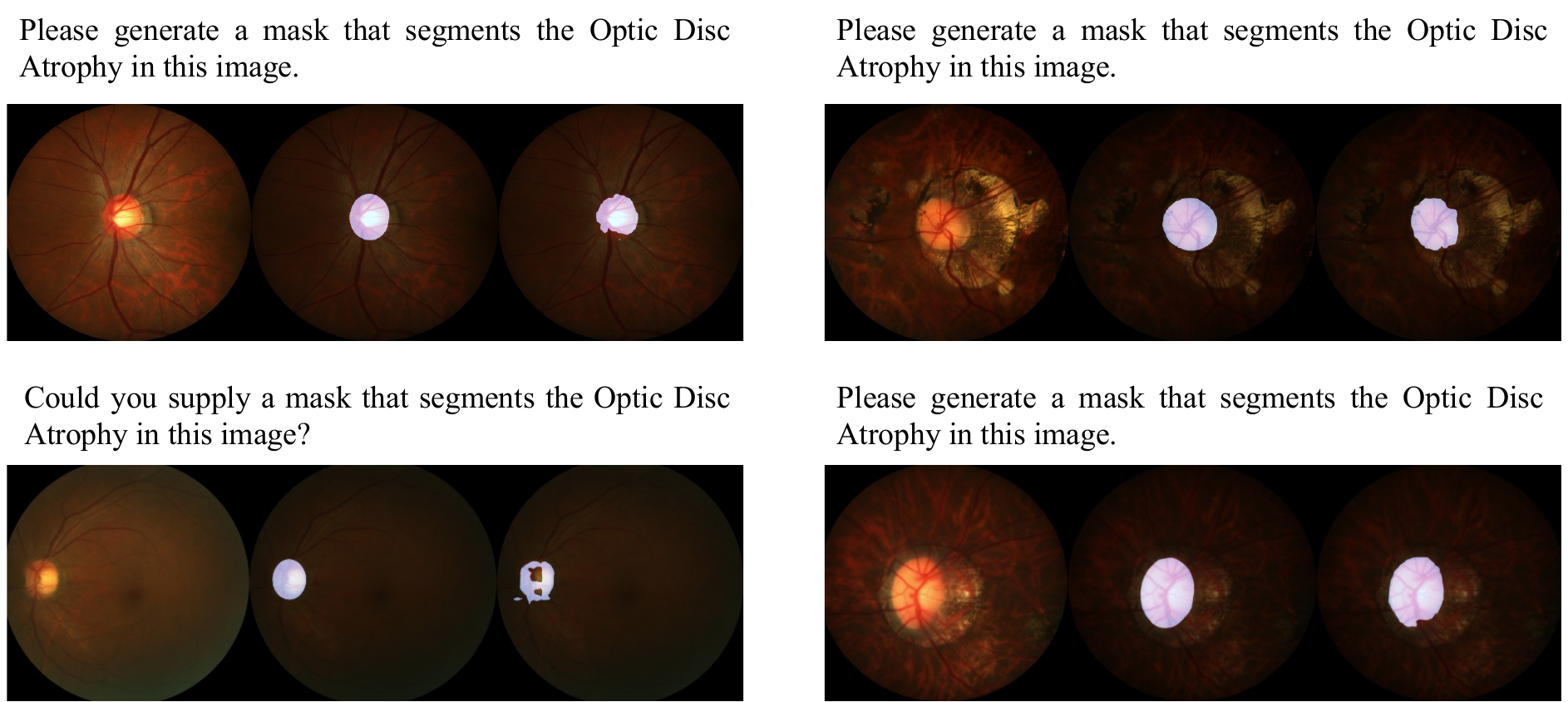}}
\caption{The display of MedPLIB's zero-shot effects in fundus modalities. The text prompt is positioned at the top, with the original input image on the left, the grounding ground truth in the center, and MedPLIB's grounding results on the right.
} \label{fig:vis_fundus}
\end{figure*}

\begin{figure*}[h]
\centerline{\includegraphics[width=1.0\textwidth]{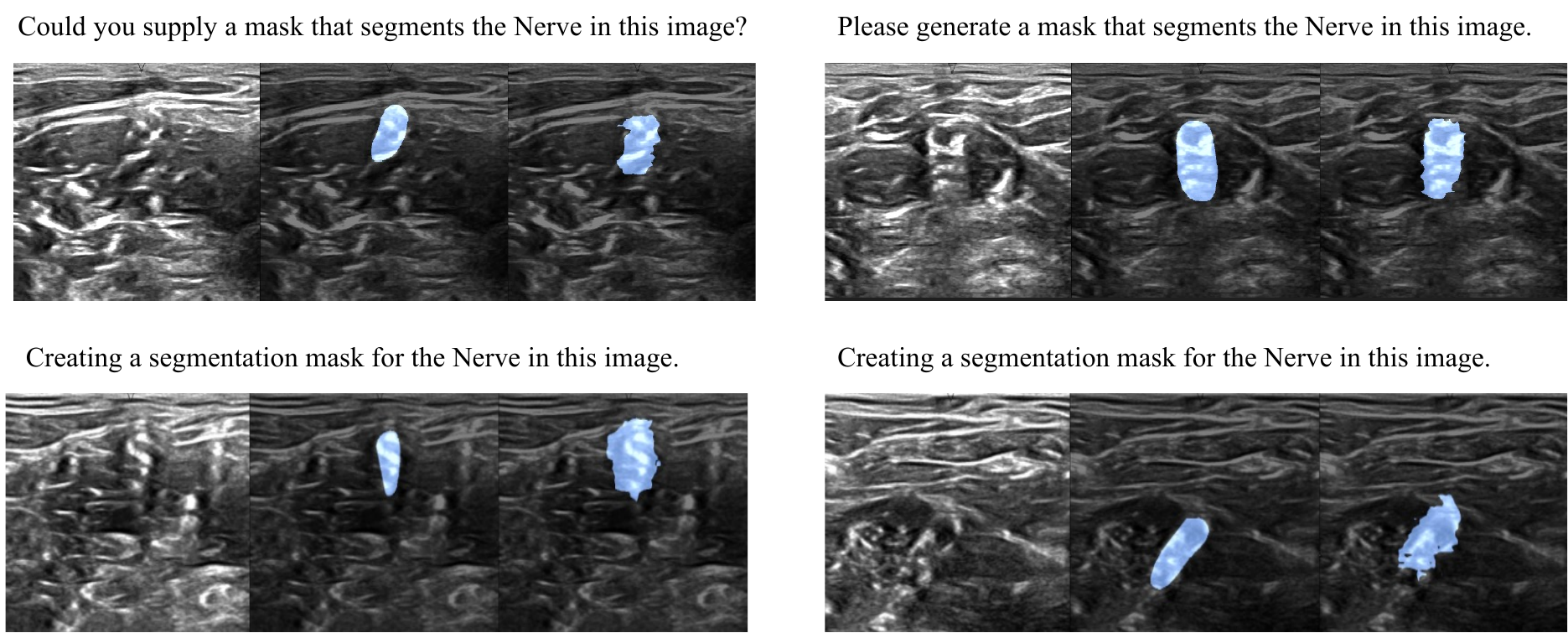}}
\caption{The display of MedPLIB's zero-shot effects in ultrasound modalities. The text prompt is positioned at the top, with the original input image on the left, the grounding ground truth in the center, and MedPLIB's grounding results on the right.
} \label{fig:vis_us}
\end{figure*}

\begin{figure*}[h]
\centerline{\includegraphics[width=1.0\textwidth]{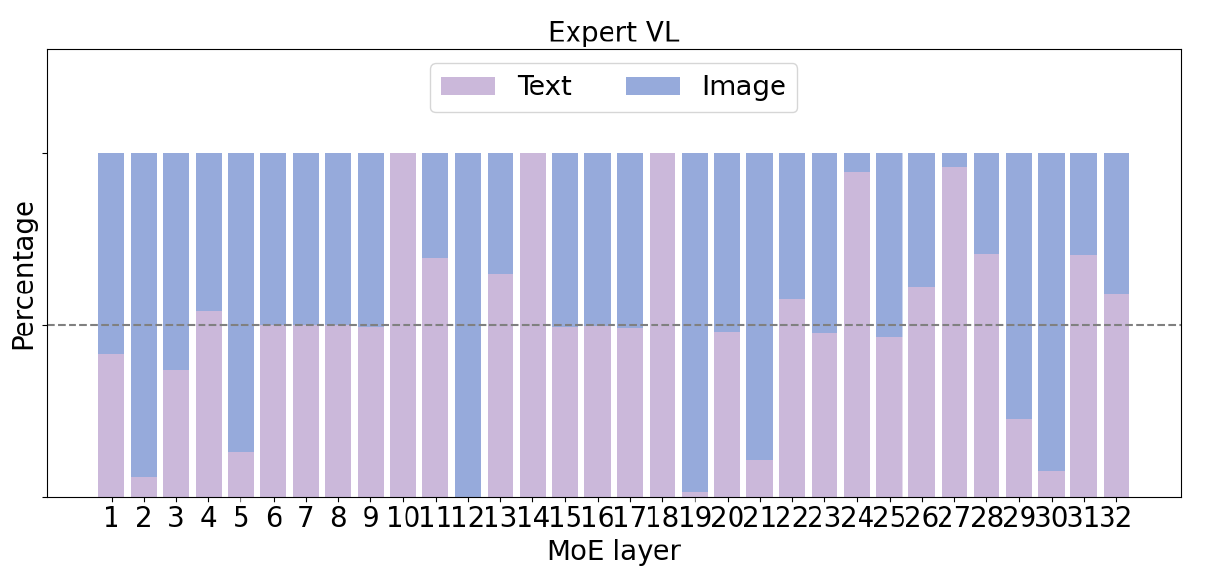}}
\caption{The preferences of expert $E_{vl}$ towards various modalities.
} \label{fig:ep_vis_vl}
\end{figure*}

\begin{figure*}[h]
\centerline{\includegraphics[width=1.0\textwidth]{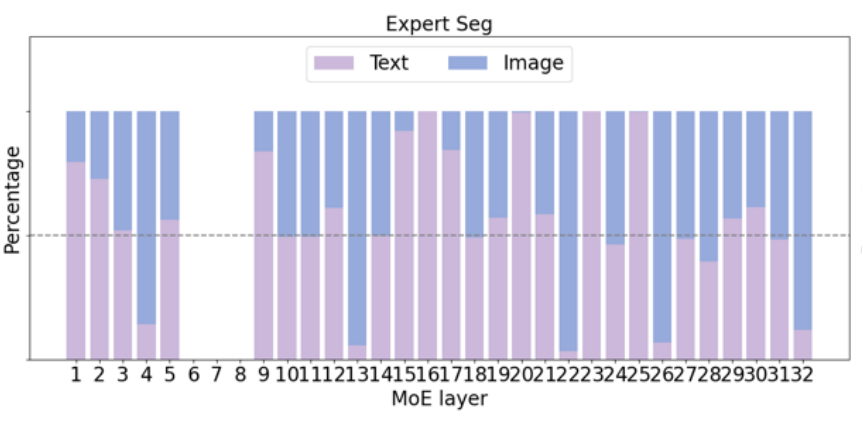}}
\caption{The preferences of expert $E_{ground}$ towards various modalities.
} \label{fig:ep_vis_seg}
\end{figure*}

Here, we visualize the results for CT (Figure~\ref{fig:vis_ct}), dermoscopy (Figure~\ref{fig:vis_der}), and PET (Figure~\ref{fig:vis_pet}) on the MeCoVQA-G test set. We also visualize the results on the fundus (Figure~\ref{fig:vis_fundus}) and ultrasound (Figure~\ref{fig:vis_us}) images during zero-shot tests. Additionally, we visualize with greater granularity the distribution of image and text information tokens processed at each layer by $E_{ground}$ (Figure~\ref{fig:ep_vis_seg}) and $E_{vl}$ (Figure~\ref{fig:ep_vis_vl}). We observed that overall, both experts maintain a relatively balanced number of image and text tokens processed at each layer. However, in specific layers, there is a tendency to focus more on either all images or all text tokens, indicating that both experts have learned a certain pattern that enables them to divide their tasks in a specific manner. 

\section{Instruction Templates}
\begin{figure*}[h]
\centerline{\includegraphics[width=1.0\textwidth]{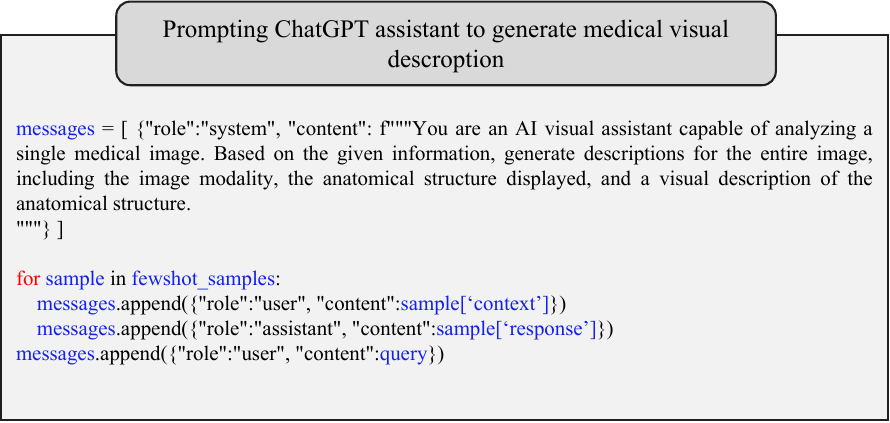}}
\caption{The template for prompting the ChatGPT to generate medical visual descriptions.
} \label{fig:instruction1}
\end{figure*}

\begin{figure*}[h]
\centerline{\includegraphics[width=1.0\textwidth]{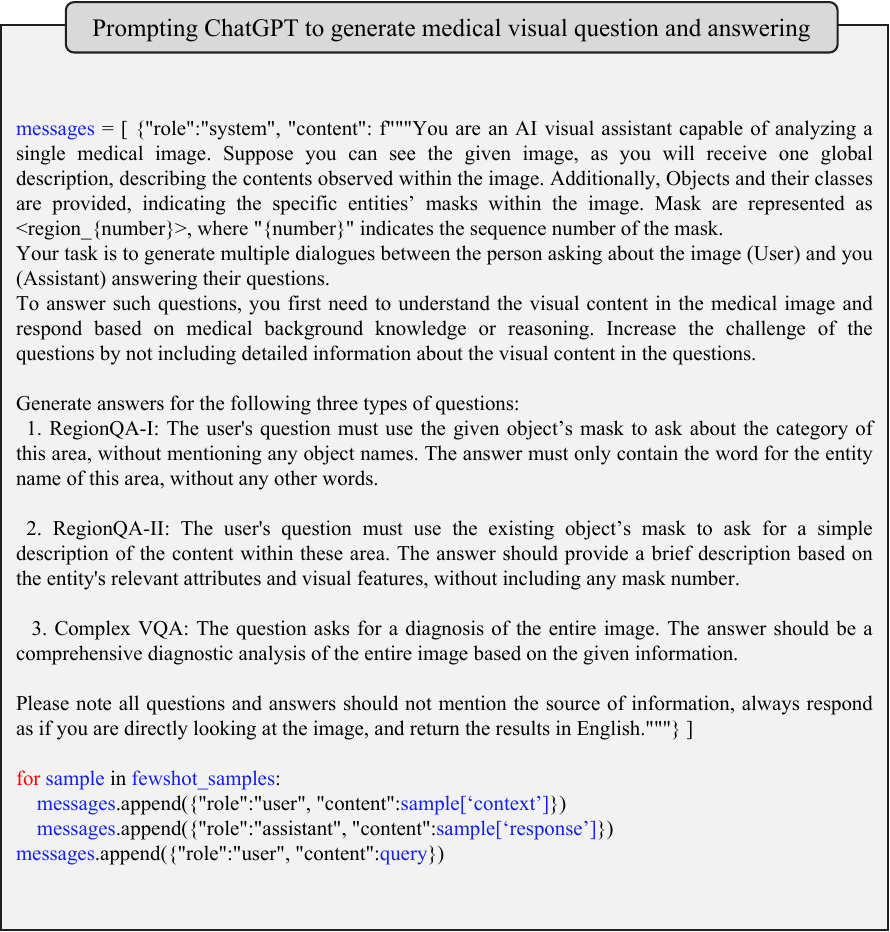}}
\caption{The template for prompting the ChatGPT to generate medical visual questions and answers.
} \label{fig:instruction2}
\end{figure*}

In Figure~\ref{fig:instruction1} and \ref{fig:instruction2}, we respectively present the pseudo-code prompts used in the first and second steps.

\section{Limitations.}
\label{sec:limitations}
While we believe that MedPLIB represents a significant step towards building a useful biomedical visual assistant, we note that it is limited by hallucinations and substantial errors in pixel-level grounding of small targets. Future work is directed toward improving both the quality and reliability of results.

\end{appendices}

\end{document}